\documentclass[acmtog, nonacm]{acmart}
\acmSubmissionID{1121}

\usepackage{booktabs} 
\renewcommand\footnotetextcopyrightpermission[1]{} 

\usepackage[ruled]{algorithm2e} 

\SetAlFnt{\small}
\SetAlCapFnt{\small}
\SetAlCapNameFnt{\small}
\SetAlCapHSkip{0pt}

\acmJournal{TOG}

\begin{document}
\title{2L3: Lifting Imperfect Generated 2D Images into Accurate 3D}

\author{Yizheng Chen}
\affiliation{%
 \institution{Zhejiang Lab}
 \country{China}}
\email{yzchen.work@gmail.com}

\author{Rengan Xie}
\affiliation{%
 \institution{Zhejiang University}
 \city{Zhejiang}
 \country{China}
}
\email{xrg1227@gmail.com}

\author{Qi Ye}
\affiliation{%
 \institution{Zhejiang University}
 \city{Zhejiang}
 \country{China}
}
\email{qi.ye@zju.edu.cn}

\author{Sen Yang}
\affiliation{%
\institution{Zhejiang Lab}
\country{China}}
\email{yangsen@zhejianglab.edu.cn}

\author{Zixuan Xie}
\affiliation{%
 \institution{Institute of Computing Technology, Chinese Academy of Sciences}
 \country{China}}
\email{xiezixuan211@mail.ucas.ac.cn}

\author{Tianxiao Chen}
\affiliation{%
 \institution{Zhejiang University}
 \country{China}
}
\email{ctx_hangzhou@163.com}

\author{Rong Li}
\affiliation{%
 \institution{Zhejiang University}
 \country{China}
}

\author{Yuchi Huo}
\affiliation{%
 \institution{Zhejiang University}
 \country{China}
}
\email{huo.yuchi.sc@gmail.com}

\renewcommand\shortauthors{Chen et al.}

\newcommand{\refFig}[1]{Figure \ref{#1}}
\newcommand{\refTab}[1]{Table \ref{#1}}
\newcommand{\refSec}[1]{Section \ref{#1}}
\newcommand{\refEq}[1]{Equation \ref{#1}}
\newcommand{\huo}[1]{
    \textcolor{orange}{Huo: {#1}}
}
\newcommand{\yq}[1]{
    \textcolor{orange}{yq: {#1}}
}
\newcommand{\Skip}[1] {
}
\newcommand{\xie}[1] {
	\textcolor{cyan}{\bfseries{XIE: {#1}}}
}
\newcommand{\todo}[1] {
	\textcolor{red}{\bfseries{TODO: {#1}}}
}

\def\ie{\emph{i.e}\onedot}

%
%


\begin{abstract}

Reconstructing 3D objects from a single image is an intriguing but challenging problem. One promising solution is to utilize multi-view (MV) 3D reconstruction to fuse generated MV images into consistent 3D objects. However, the generated images usually suffer from inconsistent lighting, misaligned geometry, and sparse views, leading to poor reconstruction quality. To cope with these problems, we present a novel 3D reconstruction framework that leverages intrinsic decomposition guidance, transient-mono prior guidance, and view augmentation to cope with the three issues, respectively. Specifically, we first leverage to decouple the shading information from the generated images to reduce the impact of inconsistent lighting; then, we introduce mono prior with view-dependent transient encoding to enhance the reconstructed normal; and finally, we design a view augmentation fusion strategy that minimizes pixel-level loss in generated sparse views and semantic loss in augmented random views, resulting in view-consistent geometry and detailed textures. Our approach, therefore, enables the integration of a pre-trained MV image generator and a neural network-based volumetric signed distance function (SDF) representation for a single image to 3D object reconstruction. We evaluate our framework on various datasets and demonstrate its superior performance in both quantitative and qualitative assessments, signifying a significant advancement in 3D object reconstruction. Compared with the latest state-of-the-art method Syncdreamer~\cite{liu2023syncdreamer}, we reduce the Chamfer Distance error by about 36\% and improve PSNR by about 30\% .

\end{abstract}

\begin{teaserfigure}
  \centering
  \includegraphics[width=1.0\linewidth]{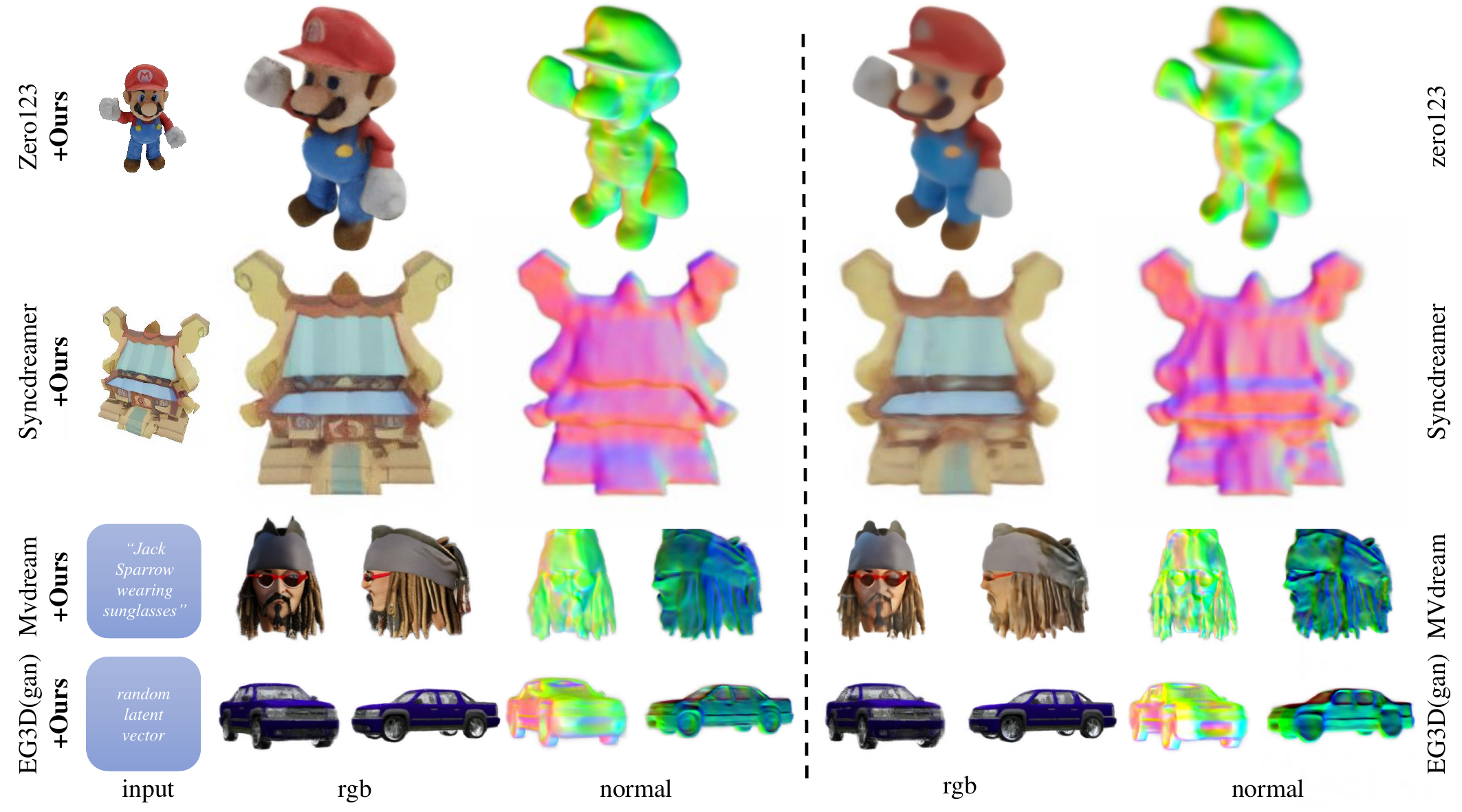}
  \caption{MeshifyDreams employs intrinsic decomposition, per-frame transient normal prior, and view augmentation to reconstruct 3D objects from generated multi-view images from a single view, text, or other inputs conditioned generation model), preserving high-quality geometry and texture.}
\label{fig:teaser}
\end{teaserfigure}
\authorsaddresses{}
\maketitle

\section{Introduction}
\label{sec:intro}

\begin{table}[ht]
\setlength{\tabcolsep}{1.0mm}
\caption{Our method enhances the quality of 3D reconstruction based on various state-of-the-art (SOTA) multi-view generators. We evaluate geometric quality with Chamfer Distance (CD) and Volume Intersection over Union (IoU), and RGB quality with PSNR, SSIM, and LPIPS, using GSO dataset for SyncDreamer and Zero123, and ShapeNet dataset for GFLA. "A+B" in the Method column represents images generated from A and 3D reconstruction using B and methods without "+" means the original implementation.}
\label{tab:generator_3d}

\begin{tabular}{lccccc}
\hline 
Method & CD $\downarrow$ & IoU $\uparrow$  & PSNR $\uparrow$ & SSIM $\uparrow$ & LPIPS $\downarrow$\\
\hline
Syncdreamer & 0.0261 & 0.542 & 18.61 & 0.722 & 0.283 \\
\textbf{Syncdreamer+Ours} & \textbf{0.0167} & \textbf{0.643} & \textbf{24.13} & \textbf{0.879} & \textbf{0.099}\\
Zero123+Neus & 0.0312 & 0.482  & 16.85 & 0.672 &0.172 \\
\textbf{Zero123+Ours} & \textbf{0.0216} & \textbf{0.585} & \textbf{21.92}& \textbf{0.714} & \textbf{0.114}\\
GFLA+Neus & 0.0527 & 0.357 & 17.93 & 0.702& 0.227 \\
\textbf{GFLA+Ours} & \textbf{0.0304} & \textbf{0.439} & \textbf{20.14} &\textbf{ 0.751} & \textbf{0.12}\\
\hline
\end{tabular}
\end{table}


Recently, there has been remarkable progress in generating 3D objects thanks to the development of generation models like diffusion and GAN \cite{rombach2022high, ramesh2022hierarchical, ho2020denoising, karras2021alias, karras2019style}. These promising results soon attracted a lot of attention and led to many intriguing possibilities, such as generating 3D objects from a single image, text prompts, or environment~\cite{poole2022dreamfusion, chan2022efficient, melas2023realfusion, wang2023prolificdreamer, lin2023magic3d}. Many of these 3D generation works leverage 2D image generation models; they usually consist of two stages: an image generation stage and a 3D reconstruction stage, and face one major challenge of reconstructing 3D objects from the generated 2D images,  \emph{i.e.} multi-view consistency.

Multi-view 3D reconstruction is a fundamental problem in 3D vision and has been extensively researched for decades. In the problem, the images are captured from real scenes, and each pixel results from the physical law of imaging: a combination of factors of light transport, object material, geometry, \emph{etc}. With these physically faithful images, 
multi-view reconstruction methods can utilize geometrical and physical priors to fuse information from multiple views and infer 3D properties like color and geometry, for example triangulating two pixels imaging from a 3D point to get its 3D position.

However, it is hard to teach a generation model to understand physical laws, and therefore hard to produce images with correct physical properties (view consistency) from generation models. To tackle the issue and utilize the 3D reconstruction techniques for 3D generation, many existing works \cite{shi2023zero123++, chan2022efficient, liu2023syncdreamer, lin2023consistent123} focus on the first stage, training or finetuing 2D image generation models with multi-views images and improving the view consistency. GFLA~\cite{ren2020deep} generates multiple views based on a GAN model and Zero123 \cite{liu2023zero} finetunes a stable diffusion model to generate a high quality novel view of the input image based on relative camera pose.
Syncdreamer \cite{liu2023syncdreamer} improves the view consistency from Zero123~\cite{liu2023zero}  via attention layers. 

\begin{figure}[!h]
  \centering
  \includegraphics[width=0.9\linewidth]{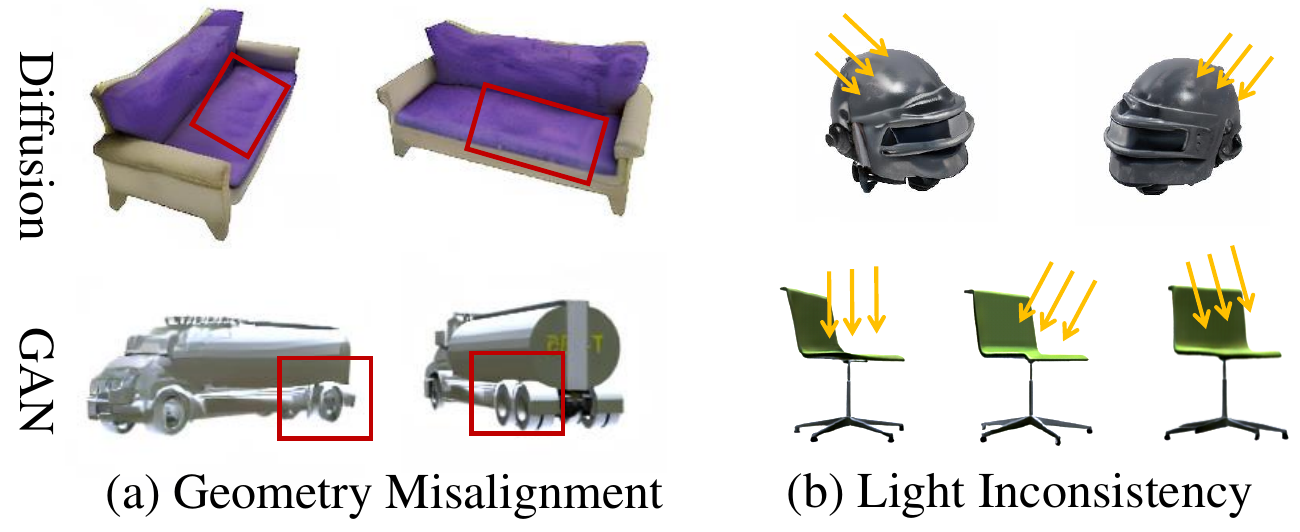}
\caption{The geometry misalignment and lighting inconsistency generally exist in state-of-the-art MV generation models like GAN~\protect\cite{chan2022efficient} and diffusion~\protect\cite{liu2023syncdreamer}. }\label{fig:abl_decomposition_intr}
\end{figure}

Though these up-to-date generation models can produce visually faithful multi-view images to some extent,  they still have a long way toward physical correctness.  For example, \refFig{fig:abl_decomposition_intr} presents a group of multi-view images generated by the state-of-the-art (SOTA) view-consistent generation techniques, including a generative adversarial network (GAN) model and diffusion model \cite{chan2022efficient, liu2023syncdreamer}. 
We notice three critical problems shared by existing multi-view generation methods: 1) \textbf{geometry misalignment}, the surfaces of the sofa and the wheels of the truck vary in different views; 2) \textbf{light inconsistency}, the important reconstruction clues like spots and shadows are view-dependent, implying an incorrect shading process that 
would significantly hinder decoupling the
geometry-shading confusion for robust geometry reconstruction; 3) \textbf{view sparsity}, because of the scarcity of 3D dataset, most generation models are trained to produce a few views without accurate camera pose. Images with these three defects significantly reduce the reconstruction quality and challenge existing 3D reconstruction techniques designed for real images.


While continuing to improve the image generator may help tackle the challenges, in the work, we instead explore the feasibility and the capability of 3D reconstruction from imperfectly generated images. We attempt to reduce the assumption of physically correct images required by current multi-view reconstruction methods and adapt the methods for the imperfect 2D images generated from off-shelf multi-view generation models. The paper presents a novel 3D reconstruction framework designed for dreamed images with the defects aforementioned. By priors learned about the 3D world from images, embedding imaging physical laws into the reconstruction, and ensuring semantic consistency between views, we alleviate the dependency of reconstruction on the physical correctness of the generated images.
Specifically,  we first introduce monocular normal prior with view-dependent transient encoding to enhance the reconstructed geometry. 
Then, we leverage to decouple the shading information from the generated images to reduce the impact of inconsistent lighting. 
Finally, we design a view augmentation fusion strategy that minimizes pixel-level loss for generated and rendered images from the same sparse views and semantic loss across different views, \emph{i.e.} generated sparse views and augmented random rendered views, resulting in view-consistent geometry and detailed textures. With our view-dependent transient encoding, decoupling of the lighting, and view augmentation, we can leverage the off-shelf models like the monocular normal estimator and decomposition models without finetuning with multi-view images, or in-domain data. This helps the generation of our method: with less dependency on finetuning models, our method can be plugged into 3D generation works based on 2D image generation directly. Also, we can easily replace the off-shelf models in our method with alternatives and keep our "plug-in" reconstruction module updated.


When replacing reconstruction modules in the existing 3D generation methods with our method, it significantly outperforms their original methods. As shown in \refTab{tab:generator_3d} (
Syncdreamer~\cite{liu2023syncdreamer} is based on 3D reconstruction with Neus\cite{wang2021neus}  and Ours is the Neus reconstruction with our proposed contributions), compared with methods with the basic Neus reconstruction (Syncdreamer, Zero123+Neus, GFLA+Neus),  methods with our reconstruction designed for dreamed images (Syncdreamer+Ours, Zero123+Ours, GFLA+Ours) gain a reduction in Chamfer Distance error about 28\% to 44\% and an improvement in PSNR about 12\% to 33\%. Also as shown in \refFig{fig:teaser}, methods with our reoncstruction attain the highest reconstruction quality, featuring smooth surfaces and intricate geometry.


In summation, our contribution includes:
\begin{itemize}
\item  We present a novel multi-view 3D reconstruction method tailored for imperfect dreamed images, which can be readily plugged into 3D generation works leveraging image generation models.
\item We discover and address the light inconsistency problem of generated images by introducing the intrinsic decomposition technique, which increases reconstruction quality and achieves the albedo component.
\item We introduce a normal prior model and per-frame transient geometry encoding to improve the geometry detail and consistency in 3D object generation.
\item We invent a view augmentation scheme to produce semantic guidance in densely sampled random views, which significantly alleviates the under-supervision problem due to sparse views. 
\end{itemize}


\section{Related work}
\label{sec:related_work}

\subsection{Single-view 3D reconstruction }

Reconstructing 3D objects from a single view is a challenging problem, as it is ill-conditioned and requires reconstructing the 3D structure of the scene from just one viewpoint.  One approach is to rely on collections of 3D primitives to approximate the target shape explicitly. These works obtain object embeddings from input RGB images and map them to the 3D space. Various 3D object representation methods are employed, such as mesh\cite{worchel2022multi, xu2019disn}, point clouds\cite{fan2017point, mescheder2019occupancy}, and voxel \cite{girdhar2016learning, wu2017marrnet}. And the methods of embedding and mapping are influenced by 3D object representation methods. Some others leverage cues like texture \cite{li2018megadepth} and defocus \cite{favaro2005geometric} to understand 3D shapes from a single image. The effectiveness of these approaches relies on the technique of estimating depth cues from images. In addition, \citet{fan2017point} directly regresses the point clouds from the image using learned priors to complete the information of invisible parts.

Recently, there has been a remarkable development of NeRF-based approaches \cite{wang2021neus,yariv2021volume} for 3D reconstruction, following the success of neural radiance fields (NeRF) \cite{mildenhall2021nerf}. Some researchers focus on improving the accuracy of sparse view reconstruction \cite{chen2021mvsnerf, chibane2021stereo, wang2021ibrnet}. Furthermore, works like PixelNeRF\cite{yu2021pixelnerf} and PVSeRF\cite{yu2022pvserf} aim to reconstruct 3D scenes from a single image by incorporating prior knowledge of the object's structure. These methods train their models on ShapeNet\cite{chang2015shapenet}, a database containing objects of simple shapes with available 3D annotation.

\subsection{Novel View Synthesis}
Novel view synthesis is the task of generating novel views of a scene from a new viewpoint given multi-view observations of the scene. The generation of high-quality images from an unseen perspective is a challenging task, particularly when the object's position and orientation in the scene are not known. One of the popular approaches to novel view synthesis involves using Generative Adversarial Network (GAN) \cite{goodfellow2020generative}. In prior works, researchers explored the use of GAN models to discover latent semantic directions that could manipulate object rotation without reliance on underlying 3D models \cite{chang2015shapenet, shen2021closed, harkonen2020ganspace}. Several recent works \cite{chan2021pi, niemeyer2021giraffe} have extended this GAN-based approach to NeRF models and trained them using adversarial losses, resulting in significant performance improvements. Auto-regressive models
have been explored \cite{sanghi2022clip, yan2022shapeformer}, which learn the distribution of these 3D shapes conditioned on images or texts.

Another promising approach to novel view synthesis involves using diffusion models, specifically diffusion-denoising probabilistic models. Diffusion models are a class of generative models that make use of a Markovian noising process to iteratively reverse the noise. In recent years, several researchers \cite{melas2023realfusion, poole2022dreamfusion,li20223ddesigner,lin2023consistent123,liu2023zero, shi2023zero123++} have explored the use of diffusion models in conjunction with radiance fields and have demonstrated excellent results in tasks such as conditional synthesis, completion, and other related tasks \cite{zeng2022lion,zhou20213d}.

\subsection{Multi-view/3D generation with 2D Generation}

2D generative models \cite{rombach2022high, ramesh2022hierarchical} have learned a wide range of visual concepts by pretraining on large-scale image datasets. They possess powerful priors about the 3D world, which allow them to exhibit significant potential in multi-view or 3D generation. Some attempts \cite{jun2023shap, nichol2022point} have been made to directly train 3D diffusion models using different 3D representations. However, this approach often require a large 3D dataset, and currently such datasets are inadequate for capturing the intricacies of diverse 3D shapes.

To leverage the capability of 2D generative models, one line of approaches \cite{poole2022dreamfusion, melas2023realfusion, wang2023prolificdreamer} propose Distillation Sampling to generate 3D assets from texts. These approaches usually suffer from low diversity, over-saturation, and multi-face problems. Some approaches attempt to generate multi-view images by a direct application of 2D diffusion models. 
Several works \cite{liu2023zero, liu2023syncdreamer, shi2023zero123++} fine-tune the StableDiffusion model \cite{saharia2022photorealistic} on a large-scale 3D render dataset. The fine-tuned models are able to generate a high-quality novel view of the input image based on relative camera pose, but the generated images still suffer from inconsistency in geometry and colors. To alleviate the issue of inconsistency, the work \cite{qian2023magic123} employs a distillation sampling approach, but the novel views are guided by a combination of 2D and 3D diffusion priors. Some other works \cite{ chan2023generative} resort to estimated depth maps to warp and in paint novel view images. However, the results heavily rely on the quality of the depth estimator; an inaccurate depth map would lead to low-quality results. 
\cite{chan2023generative, tewari2023diffusion} generate new images using an autoregressive render-and-generate approach, but limited to specific object categories or scenes. Recent work \cite{liu2023syncdreamer, szymanowicz2023viewset} produce consistent multiview color images via attention layers but face low-quality geometry and blurring textures challenges.

\section{Proposed Framework}
\label{sec:propesed_framework}

\begin{figure*}[!h]
  \centering
  \includegraphics[width=\linewidth]{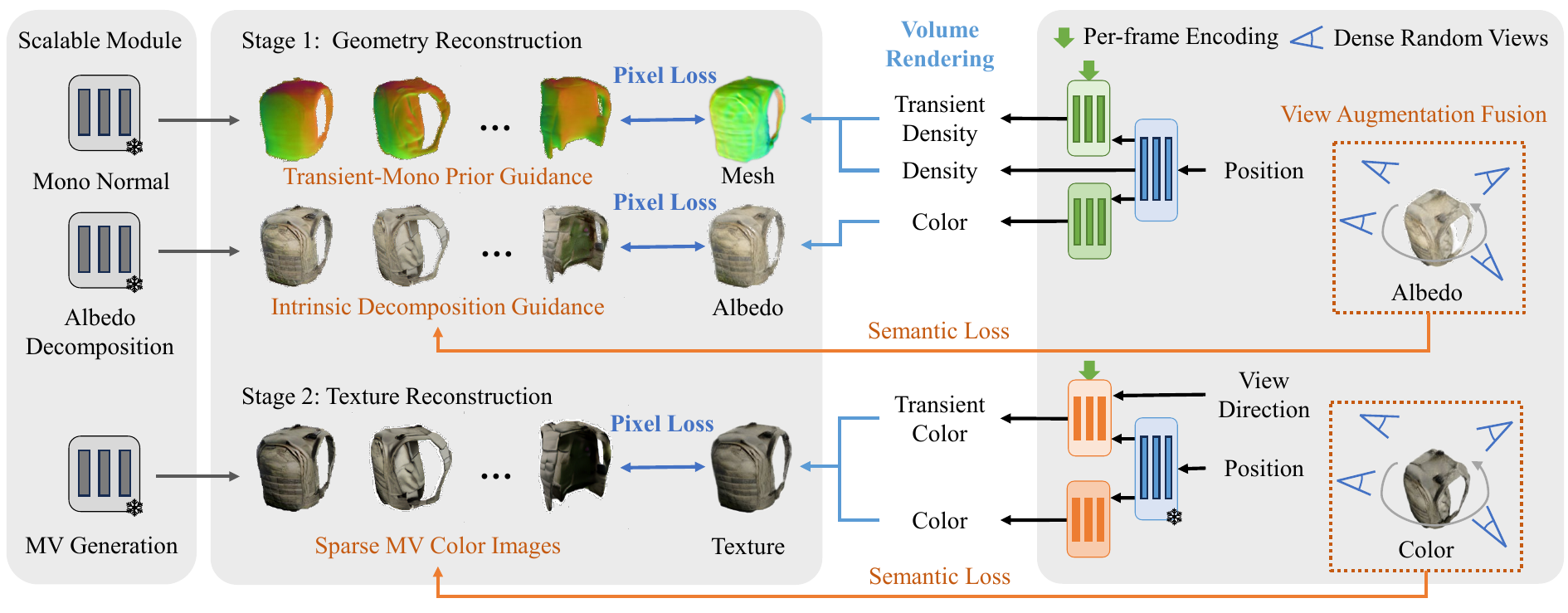}
  \caption{Our pipeline of 3D mesh reconstruction from generated multi-view images. Off-shelf models for 2D images generation, instrinsic decomposition and monocular depth estimation are leveraged to generate sparse multi-view images, and their normal and albedo maps for supervision in the reconstruction stages. Our reconstruction is decomposed into two stages to produce view-consistency 3D results. Stage 1: reconstructing the geometry and albedo field with the guidance of normal and albedo maps. Stage 2: reconstructing shaded texture with highlight and shadow details. Further, per-frame encoding and view augmentation fusion schema are designed to enhance view consistency and alleviate under-supervision of sparse views.}\label{fig:mainfigure}
\end{figure*}

\subsection{Overview}
\refFig{fig:mainfigure} illustrates our pipeline. It takes a single image or other input conditions (\refFig{fig:mainfigure} only shows image for example) as input and uses pre-trained MV generation models to generate sparse images for 3D reconstruction via training a neural SDF-based representation. The reconstruction can be roughly classified into two stages.

The first stage reconstructs the geometry and albedo field. We utilize a pretrained mono-depth prediction model to synthesize normal prior of the spare images as geometry guidance. Considering that the geometry is view-inconsistent, we further integrate a per-frame encoding to predict the transient part of geometry in each view. Regarding the lighting, we use a pretrained intrinsic decomposition model to decompose the image and use its albedo for 3D reconstruction. This scheme can eliminate the inconsistent shading clues, e.g., the specular light spots, in the generated images, thus improving the reconstructed geometry. Furthermore, we can directly achieve the albedo component for downstream applications like relighting. Finally, we invent a view augmentation scheme that densely samples tens of random views to enrich the sparse views. Because there are no ground-truth images on these random views, we minimize the semantic loss between the rendered images and its nearby sparse views.  

The second stage reconstructs the texture, producing a shaded texture with highlight and shadow details. The process is similar to the first stage. Specifically, we freeze the geometry and albedo fields reconstructed by the first stage, and use the generated sparse images to reconstruct an RGB field for texturing the mesh.

\subsection{Neural Volume Rendering}

Following NeuS \cite{wang2021neus}, we optimize the implicit SDF field and color field to reconstruct the 3D object using volume rendering. Given a pixel, we denote the ray emitted from this pixel as $\{\mathbf{p}(t)=\mathbf{o}+t \mathbf{v} \mid t \geq 0\}$, where $\mathbf{o}$ is the center of the camera and $\mathbf{v}$ is the unit direction vector of the ray. We accumulate the colors along the ray by
\begin{equation}
C(\mathbf{o}, \mathbf{v})=\int_0^{+\infty} w(t) c(\mathbf{p}(t), \mathbf{v}) \mathrm{d} t,
\end{equation}
where $c(\cdot)$ denotes an implicit color field represented by a neural network.  The weighting function $w(t)=T(t)\rho(t)$. $T(t)$ is computed as: $T(t)=\exp \left(-\int_0^t \rho(u) \mathrm{d} u\right),$ and $\rho(t)$ is opaque density defined as:
\begin{equation}
\rho(t)=\max \left(\frac{-\frac{\mathrm{d} \Phi_s}{\mathrm{~d} t}(f(\mathbf{p}(t)))}{\Phi_s(f(\mathbf{p}(t)))}, 0\right),
\end{equation}
where $\Phi_s(\cdot)$ is the Sigmoid function and $f(\cdot)$ is the SDF value retrieved from a neural network-based geometry representation.






\subsection{Geometry Reconstruction Stage}
In contrast to conventional geometry reconstruction that aims to faithfully match the input content, we need to take the inconsistency between generated images into consideration. Therefore, we employ a combination of techniques, including the utilization of mono prior, per-frame normal encoding, and intrinsic decompose, all of which play an important role in the optimization process.


\paragraph{Intrinsic Decomposition Guidance.}
Intrinsic image decomposition (IID) is the process of recovering the image formation components, such as reflectance (albedo) and shading (illumination) from an image. Here, we employ IID to separate material properties and shading information from input images to reduce the impact of inconsistent lighting, and then use the separated albedo to reconstruct the color field in the geometry reconstruction stage.


Given a generated image $I$, we decomposed it using Pie-net~\cite{das2022pie} as the pixel-wise product of the albedo $I^{albedo}$ and the illumination variance $I^{shade}$ as
\begin{equation}
I= I^{\text{albedo}} \odot I^{\text{shade}},
\end{equation}
and minimize the difference between rendered pixel color $\hat{C}$ and the pixel color of $I^{\text{albedo}}$ in Stage 1.

\paragraph{Mono Normal Prior.} Because the object boundary in 2D images primarily shapes the object's 3D contour, it requires dense views to determine geometry details. However, it is hard to achieve a lot of images via existing generation models. Furthermore, generating more images might not help because the generated geometry is misaligned from different views and produces conflicts. To address this problem, we employ readily available monocular geometric priors, Omnidata~\cite{eftekhar2021omnidata}, to generate a normal map $\bar{N}$ for each RGB image. Unlike depth cues, which provide semi-local and relative information, normal cues offer localized insights into the geometric structure of the scene. 

\begin{figure}[!h]
  \centering
  \includegraphics[width=0.90\linewidth]{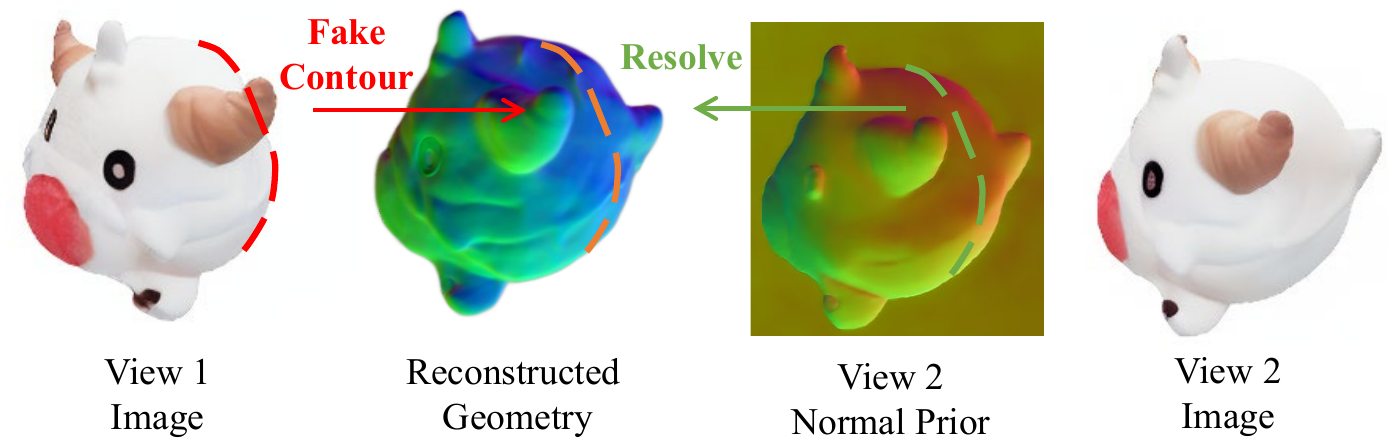}
  \caption{The red line of view 1 represents a misaligned boundary in view 2, which might lead to a wrong contour on the surface of view 2, as shown by the orange line. However, the mono normal prior of view 2 enforces a smooth constraint on the same region (the green line), and thus eliminates the wrong contour in the final reconstructed geometry.}\label{fig:contour}
\end{figure}

In addition, the fake contour due to geometry misalignment can be resolved by mono prior, as illustrated in \refFig{fig:contour}.

\paragraph{Per-Frame Normal Encoding.} In 3D reconstruction, monocular normals hold reference value, yet inconsistencies arise due to the disparate perspectives of monocular estimations. Therefore, we integrate per-frame encoding to decouple the transient part of each image.

To allow the transient component of the scene to vary across images, we assign each training image $I$ a second embed$\operatorname{ding} \ell^{(n)} \in \mathbb{R}^{n^{(n)}}$, which is given as input to one branch of the SDF network  as show in the Stage 1 in \refFig{fig:mainfigure}. Here we abuse the notation a bit for brevity and use $f(\mathbf{p}(t), \ell^{(n)})$ to denote the SDF network consisting of both the transient branch and the non-transient branch. Therefore the SDF value of a point is given by $f(\mathbf{p}(t), \ell^{(n)})$. We denote the surface normal at the point as $\nabla f(\mathbf{p}(t), \ell^{(n)})$. 


To render the normal maps, we follow the methodology in NeRF \cite{mildenhall2021nerf} and NeuS \cite{wang2021neus}. This scheme samples $n$ points $\left\{\mathbf{p}_i=\mathbf{o}+t_i \mathbf{v} \mid i=\right.$ $\left.1, \ldots, n, t_i<t_{i+1}\right\}$ along the ray to compute the approximate normal of the ray as
\begin{equation}
\hat{N}=\sum_{i=1}^n T_i \alpha_i \nabla f(\mathbf{p}_i, \ell^{(n)}).
\end{equation}

We impose consistency on the volume-rendered normal $\hat{N}$ and the predicted monocular normal $\bar{N}$ transformed to the same coordinate system with angular and L1 losses
\cite{eftekhar2021omnidata}:
\begin{equation}
\mathcal{L}_{\text {norm}}=\sum_{\mathbf{r} \in I}\|\hat{N}(\mathbf{r})-\bar{N}(\mathbf{r})\|_1+\left\|1-\hat{N}(\mathbf{r})^{\top} \bar{N}(\mathbf{r})\right\|_1.
\end{equation}

\subsection{Texture Reconstruction}

In the second stage, we use the trained geometry field to infer density while training a texture field that aims to faithfully represent the generated image $I$.

\textbf{Per-frame Color Encoding.}
Because the generated image $I$ also contains many inconsistent details, we integrate a per-frame color encoding like the transient per-frame normal encoding.
Similar to fer-frame normal encoding, we add $\ell^{(c)}  \in \mathbb{R}^{n^{(c)}}$ into a transient color network:
\begin{equation}
c_i^{(\tau)}=M\left(\mathbf{p}_i, \boldsymbol{v}_{p_i}, \nabla f(\mathbf{p}_i, \ell^{(n)}), \boldsymbol{z}_{p_i}, \ell^{(c)} \right),
\label{eq:color}
\end{equation}

\begin{equation}
\hat{C}^{(\tau)}=\sum_{i=1}^n T_i \alpha_i c_i^{(\tau)}.
\end{equation}



The transient color network $M$ is a neural network (MLP) that takes into account the surface point $\mathbf{p}_i$, the viewing direction $\boldsymbol{v}_{p_i}$, the corresponding surface normal $\nabla f(\mathbf{p}_i, \ell^{(n)})$, the latent feature vector for the point ${\boldsymbol{z}}_{p_i}$ from Stage 1, and the per-frame color encoding $\ell^{(c)}$. 

The transient color and the non-transient color are combined to form the rendered color $\hat{C}(r)$ for a ray $r$ as shown in the bottom-left of \refFig{fig:mainfigure}

\Skip{
\textbf{Hash Encoding.}
To overcome the slow training time of deep
coordinate-based MLPs, which is also a main reason for the
slow performance of NeuS, recently, Instant-NGP \huo{cite} proposed a multi-resolution hash encoding and has proven its
effectiveness. Specifically, Instant-NGP assumes that the
object to be reconstructed is bounded in multi-resolution
voxel grids. The voxel grids at each resolution are mapped
to a hash table with a fixed-size array of learnable feature
vectors assigned at the surrounding voxel grids at this level. Besides the hash encoding, another key factor to
the training acceleration is the CUDA implementation of
the whole system, which makes use of GPU parallelism.
While the runtime is significantly improved, Instant-NGP
still does not reach the quality of NeuS in terms of geometry reconstruction accuracy.
}

\subsection{View Augmentation Fusion}
Our goal is to reconstruct a 3D model from sparse multi-view images $\{I_{s}\}_{s=0}^{N}$. Intuitively, we can train an SDF field directly from  $\{I_{s}\}$ utilizing volume rendering. However, a set of sparse generated images lacks supervision from many viewpoints, leading to issues with the generated texture that may be unreasonable or unclear. To address this issue, we propose a view augmentation fusion that aims to provide supervision from any viewpoint, resulting in texture with a high degree of fidelity. Specifically, there are two main points to this strategy:

\paragraph{Asymmetric Pixel-level Loss.} We adopt a selective optimization approach that focuses on optimizing pixel-wise mean squared error (MSE) losses only in a limited set of generated images that may exhibit overlap with each other. We adopt an asymmetric pixel-level RGB loss to minimize the per-pixel difference between the predicted view and the generated images $I_{s}$. The pixel-level RGB loss is

\begin{equation}
\mathcal{L}_{rgb} = w_s\sum_{r}^R ||\hat{C}(r) - C(r)||_2^2 
\label{eq:colorloss},
\end{equation}
where $r$ denotes a ray sampled from a $s^{th}$  image of a training image pool $\{I_{s}\}_{s=0}^{N}$ consisting of $N$ generated sparse views and one input view, and $R$ denotes a set of sampled rays.  $\hat{C}(r)$ is the predicted pixel color at ray $r$ and $C(r)$ is the pixel color in reference images. In Stage 1, the reference images are albedo maps $\{I_{s}^{albedo}\}$ decomposed from $\{I_{s}\}$ and in Stage 2, the original sparse multi-view images $\{I_{s}\}$. 
$w_s$ is the weight of loss for the $s^{th}$  image, which reflects the credibility of different views. In our work, we apply $w_s=|v_s-v_0|$, where we assume the input single view $v_0$ to be the origin view.
This approach helps to minimize the impact of pixel-level misalignment and generate sharper training results with reduced blurring.

\paragraph{Semantic Consistency.} We propose to employ a self-supervised semantic loss to connect the striding key viewpoints to further enhance view consistency across the generated images. This involves incorporating a pre-trained Vision Transformer (ViT) network, which has been proven to be an expressive semantic prior even between images with misalignment \cite{tumanyan2022splicing, amir2021deep}.
 Inspired by \cite{jain2021putting}, for each image $I_s$ we randomly sample $J$ unseen viewpoints $\left\{p_j\right\}_{j=1}^{J}$ around the object and render the images $\left\{I_{p_j}\right\}_{j=1}^{J}$ from the SDF field utilizing volumetric rendering. Furthermore, we adopt a pre-trained ViT model $E_{vit}$ to extract feature embedding from images and enforce semantic consistency by minimizing the difference of feature embedding between different views: 
\begin{equation}
\begin{split}
\mathcal{L}_{sem} = & \sum_{j=1}^J ||E_{vit}(I_0) - E_{vit}(I_{p_j})||_2^2  \\
+ &\sum_{s=1}^N \sum_{j=1}^J w_{sj}||E_{vit}(I_s) - E_{vit}(I_{p_j})||_2^2,
\label{eq:semloss}
\end{split}
\end{equation}
where $E_{vit}(I_0)$, $E_{vit}(I_s)$, and $E_{vit}(I_{p_j})$ are the semantic features of the input image, generated multi-view images, and the rendered image from a random viewpoint, respectively. In Stage 1, these features are extracted from corresponding albedo maps. $w_{sj}=|v_s-p_j|$. This term compares the semantic features of the reference images and the rendered images from any random viewpoint to ensure the consistency of the underlying scene structure. In practice, we adopt CLIP-ViT \cite{radford2021learning}, a self-supervised vision transformer trained on ImageNet \cite{deng2009imagenet} dataset.

We apply Asymmetric Pixel-level Loss and Semantic Consistency Loss to both reconstruction stages. 
\subsection{Loss Design}
\Skip{
\paragraph{Pixel Loss} 

We adapt the pixel-level RGB loss from NeRF to minimize the per-pixel difference between the predicted view and the pseudo views. Let $\mathbf{c}(x,y)$ \huo{better to use other symbols rather than x,y since they have been defined earlier } be the predicted RGB color at pixel $(x,y)$ and $\mathbf{q}(x,y)$ \huo{q or C?} be the ground truth RGB color at the same pixel. Then, the pixel-level RGB loss is defined as:
\begin{equation}
\mathcal{L}_{rgb} = \sum_{i=1}^N ||\mathbf{C}(x,y) - \bar{C}(x,y)||_2^2,
\end{equation}
where $\mathbf{C}(x_i)$ is the predicted pixel color at position $x_i$, $\mathbf{C}_i$ is the  pixel color in pseudo views, and $N$ is the total number of pixels in the image.
}

\paragraph{Eikonal Loss.}
Following common practice \cite{yariv2021volume}, we also add an Eikonal term \cite{gropp2020implicit} on the sampled points $\mathcal{X}$ to regularize SDF values in 3D space:
\begin{equation}
\mathcal{L}_{eik}=\sum_{\mathbf{x} \in \mathcal{X}}\left(\left\|\nabla f_\theta(\mathbf{x})\right\|_2-1\right)^2.
\end{equation}

The combined loss function is given by:
\begin{equation}
\mathcal{L} =  \mathcal{L}_{rgb} +  \mathcal{L}_{sem} + \lambda_1 \mathcal{L}_{eik}
 + \lambda_2 \mathcal{L}_{norm},
\end{equation}
where $\lambda$s are hyperparameters that control the relative importance of each loss term.
\section{Experiments Setup}
\label{sec:experiments_details}

In order to evaluate the proposed reconstruction framework, we leverage a series of pre-trained models. Regarding MV image generation, we use the SOTA diffusion model presented by SyncDreamer~\cite{liu2023syncdreamer} and the SOTA GAN model EG3D~\cite{chan2022efficient}. Besides, we use Omnidata~\cite{eftekhar2021omnidata} for monocular normal prior generation and Pie-net~\cite{das2022pie} for intrinsic image decomposition.

\subsection{Implementation Details}
Our proposed method is trained on a single NVIDIA A40. During training, we use Adam optimizer with a learning rate of 1e-4 and a batch size of 32. We typically take 50k iterations for the SDF field. In addition, we sample 1024 ray directions each iteration when training SDF field $f_{\theta}$. The entire process takes approximately 40 minutes to train an SDF field $f_{\theta}$ of each object. 
When training the SDF field, we set $R_{\min = 16}, R_{\max =2048}$ , $L = 16$ and $n^{(n)} = n^{(c)} = 8$.\Skip{ We modify the loss weights depending on the type of training view. When the training view is a random view, we do not use $\mathcal{L}_{rgb}$, i.e., $\lambda_1 = 0$. When the training view is within $V_{key}$, we set $\lambda_1 = 0.5$. Finally, when the training view is the front view, we set $\lambda_1 = 1$.} In addition, we set $\lambda_1 =0.7$ and $\lambda_2 = 0.1 $. 

\subsection{Evaluation Metrics}

We evaluate the performance of our method on the GSO \cite{downs2022google} and report the quantitative and qualitative results in the following section. We used three standard evaluation metrics to quantitatively evaluate the performance of our proposed framework for texture reconstruction: Peak Signal-to-Noise Ratio (PSNR), Structural Similarity Index (SSIM) \cite{wang2004image}, and Learned Perceptual Image Patch Similarity (LPIPS) \cite{zhang2018unreasonable}. In addition, we employ Chamfer Distance(CD) and Volume IoU on the GSO dataset for geometry quality evaluation. 

Note that although we evaluate quantitative metrics on GSO, the following qualitative results include many out-of-domain images, such as real-world photos, 2D design images, and pictures from the internet. This is done to showcase the generalization and robustness of our work.
\Skip{PSNR measures the quality of the reconstructed 3D objects in terms of signal-to-noise ratio, while SSIM measures the structural similarity between the reconstructed 3D objects and ground truth. LPIPS is a perceptual distance metric that is learned from a deep neural network and provides a more accurate evaluation of the visual quality of the reconstructed results.}

\begin{figure}[!h]
\centering
\includegraphics[width=1.0\linewidth]{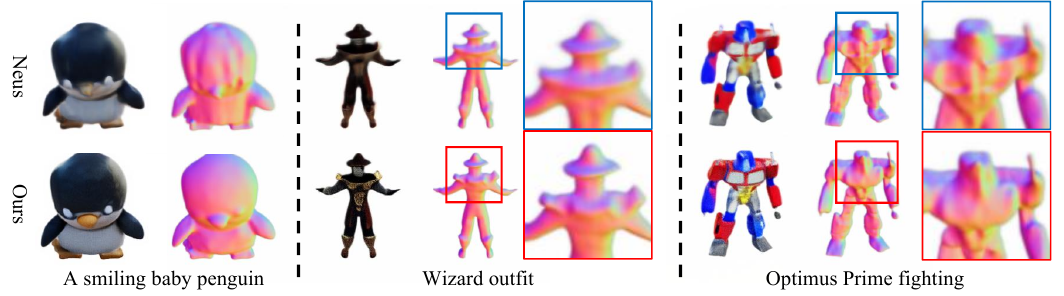}
  
  \caption{Visual comparison of reconstruction using baseline and our framework on text-generated images~\protect\cite{shi2023mvdream}. Our method can be extended to multi-view images produced using various methods based on different inputs.}
   \label{fig:gan}
\end{figure}

\begin{table}[h]
\setlength{\tabcolsep}{3.2mm}
\caption{Quantitative comparison with SOTA 3D generation methods. We report Chamfer Distance and Volume IoU on the GSO dataset.}
\label{tab:3D_results}
\begin{tabular}{ccc}
\hline 
Method & CD $\downarrow$ & IoU $\uparrow$ \\
\hline
Realfusion \cite{melas2023realfusion} & 0.0819 & 0.274 \\
Magic123 \cite{qian2023magic123} & 0.0516 & 0.453 \\
One-2-3-45 \cite{liu2023one} & 0.0629 & 0.409 \\
Point-E \cite{nichol2022point} & 0.0426 & 0.288 \\
Shap-E \cite{jun2023shap} & 0.0436 & 0.358 \\
\textbf{Zero123 + Ours} & \textbf{0.0216} & \textbf{0.585} \\
\textbf{SyncDreamer + Ours}  & \textbf{0.0167}& \textbf{0.643} \\
\hline
\end{tabular}
\end{table}

\subsection{Application on Various Multi-View Generators}
As discussion in~\refSec{sec:intro}, the~\refTab{tab:generator_3d} and~\refFig{fig:teaser} show that our method significantly improves the quality of 3D reconstruction when applied to a variety of multi-view generators. In addition, as shown in the quantitative results in~\refTab{tab:3D_results}, when our method is applied to Zero123~\cite{liu2023zero} and SyncDreamer~\cite{liu2023syncdreamer}, it consistently achieves outcomes that surpass those of the current state-of-the-art methods.

Furthermore, we present the qualitative comparison in~\refFig{fig:compare}. It is clear that Shap-E~\cite{jun2023shap} often generates incomplete meshes, failing to reconstruct regions with rich geometric details, which results in mesh holes. The One-2-3-45~\cite{liu2023one} model employs a 3D convolutional network and feature volume to extract spatial information from the inconsistent multiview outputs of Zero123 and uses an MLP for direct SDF prediction, speeding up 3D reconstruction. Nonetheless, this approach often results in smoother outputs with diminished geometric and textural detail. 

In contrast, when our method is applied to the SyncDreamer~\cite{liu2023syncdreamer}, the resulting 3D reconstructions exhibit precise features, such as detailed backpack surfaces and rich textural details like the feathers of a bird, surpassing all other state-of-the-art methods in quality. Our approach is also applicable to multi-view images generated by other methods, as shown in~\refFig{fig:gan}, demonstrating the distinction between directly using Neus~\cite{wang2021neus} and using our method reconstruction on text-generated~\cite{shi2023mvdream} images. In fact, our method is applicable to all reconstructions from multi-view images where inconsistencies are present.


\subsection{Ablation Study}

\begin{table}[ht]
\setlength{\tabcolsep}{0.9mm}
\caption{Quantitative evaluation results for ablation study. The 2D metrics (PSNR, SSIM, LPIPS) are tested on 60 random novel views rendered by the reconstructed color field. }
\label{tab:ablation}

\begin{tabular}{lccccc}
\hline 
Method & CD $\downarrow$ & IoU $\uparrow$  & PSNR $\uparrow$ & SSIM $\uparrow$ & LPIPS $\downarrow$\\
\hline
w/o Decomposition    & 0.0211 & 0.587 & 22.52& 0.852 & 0.117\\
w/o Transient prior & 0.0214& 0.603 & 23.82 & 0.859 & 0.104\\
w/o Per-frame encoding & 0.0188 & 0.622 & 24.06 & 0.861 & 0.105\\
w/o Augmentation &0.172 & 0.640 &  21.63 &  0.791 & 0.142 \\
\textbf{Ours} & \textbf{0.0167} & \textbf{0.643} & \textbf{24.25} &\textbf{0.863} & \textbf{0.097}\\
\hline
\end{tabular}
\end{table}

\begin{figure}[!h]
  \centering
  \includegraphics[width=1.0\linewidth]{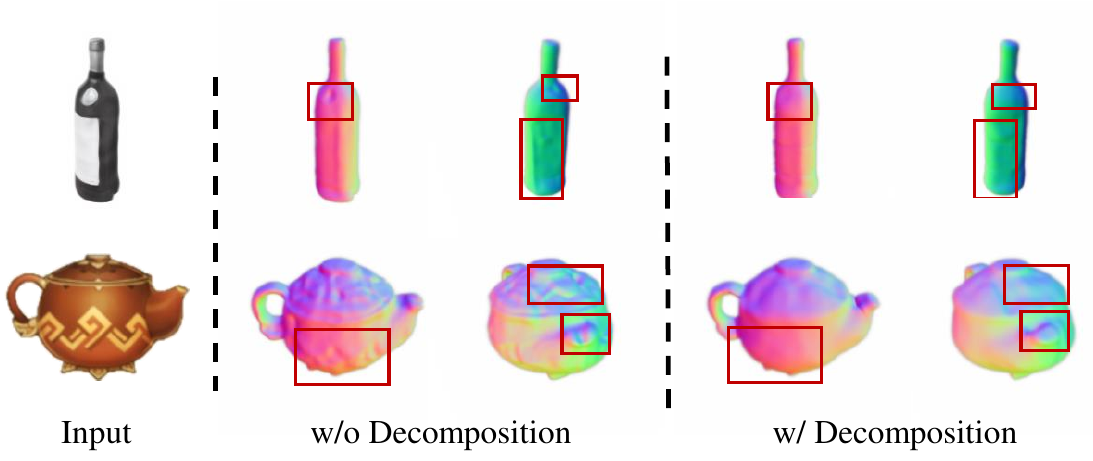}
  \caption{Visual comparison of without and with the intrinsic decomposition guidance in stage one. }\label{fig:abl_decomposition}
\end{figure}

\begin{figure}[!h]
\centering
\includegraphics[width=1.0\linewidth]{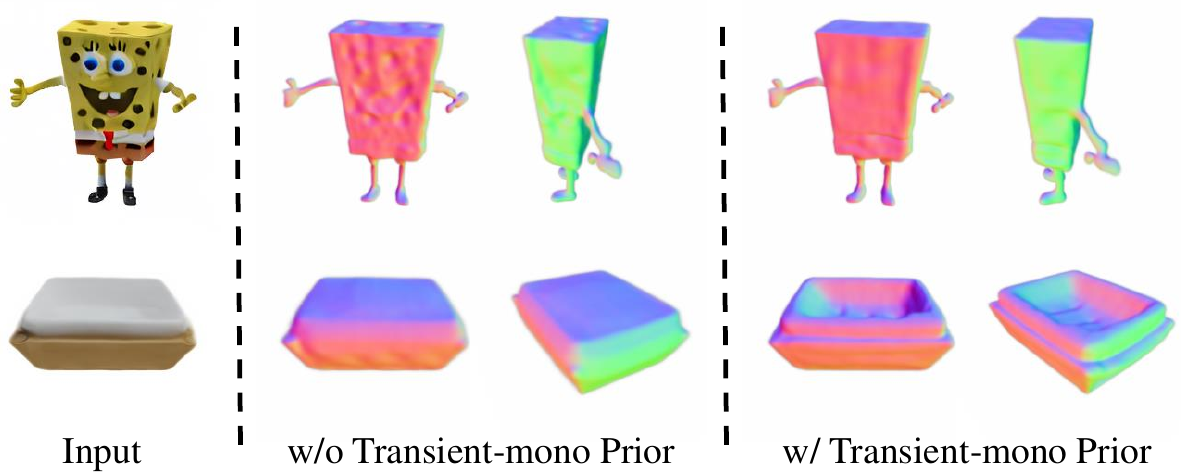}
  \caption{Visual comparison of without and with the transient-mono prior guidance in stage one. The guidance helps reconstruct geometry details and remove conflicts. }
   \label{fig:abl_normal}
\end{figure}

\begin{figure}[!h]
  \centering
  \includegraphics[width=1.0\linewidth]{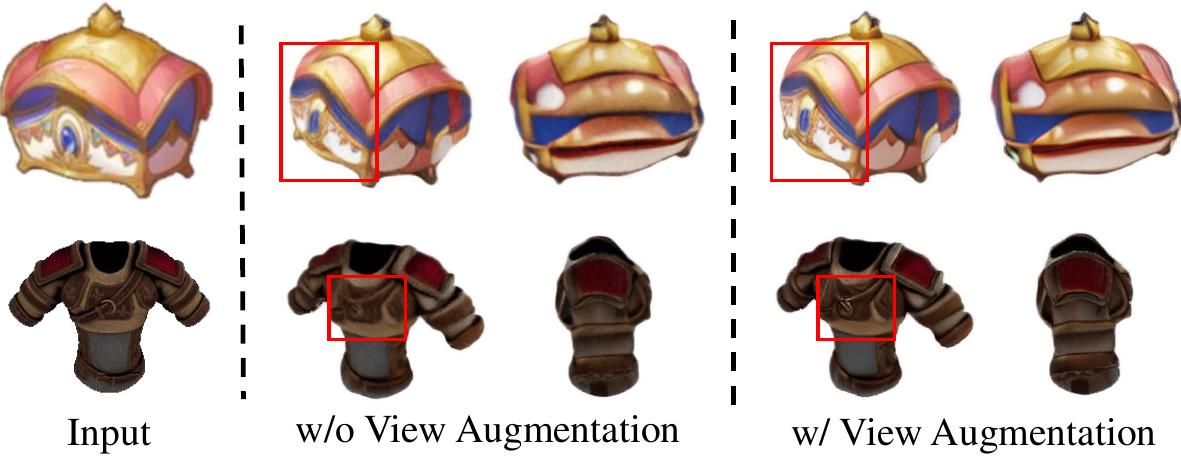}
  \caption{Visual comparison of without and with the view augmentation fusion. This strategy benefits the reconstructed texture details and visual quality.}\label{fig:abl_augmentation}
\end{figure}

In our ablation study, we systematically evaluate the contributions of different components within our framework.  The~\refFig{fig:abl_decomposition} demonstrates that the intrinsic decomposition guidance can eliminate ambiguities on the surface of a wine bottle caused by specular highlights. Without the intrinsic decomposition guidance, areas with highlights would incorrectly appear as indentations. The transient monocular normal prior significantly assists in overcoming the supervision deficit caused by sparse viewpoints, as shown in~\ref{fig:abl_normal}. This issue is particularly pronounced in regions with either overly complex or overly simplistic textures, which can lead to inaccuracies in geometric reconstruction. In addition, with augmented view supervision, the visual quality of the rendered views is improved substantially in texture detail shown as \refFig{fig:abl_augmentation}. Furthermore, we quantitatively evaluate the impact of each component of our framework, as shown in \label{tab:ablation}. Each component contributes to performance enhancement, with the complete framework delivering the best reconstruction quality, both in terms of geometry and texture.

\Skip{focusing on the intrinsic decomposition guidance~(\refFig{fig:abl_decomposition}), transient-mono prior guidance~(\refFig{fig:abl_normal}),  and view augmentation fusion~(\refFig{fig:abl_augmentation}).}



\section{Conclusion}

Our paper introduces a framework designed to reconstruct 3D representations from imperfect 2D images created by off-the-shelf multi-view generation models. Leveraging our view-dependent transient encoding, along with the decoupling of lighting and view augmentation, we employ models like the monocular normal estimator and decomposition models without the need for finetuning with multi-view images or in-domain data. This approach allows our framework to be easily integrated into 3D generation works that are based directly on 2D image generation. Additionally, the flexibility of our method means that the off-shelf models can be readily replaced with alternatives, keeping our scalable reconstruction module up-to-date.

\Skip{
This paper presents our novel framework, This helps the generation of our method: with less dependency on finetuning models, our method can be plugged into 3D generation works based on 2D image generation directly. Also, we can easily replace the off-shelf models in our method with alternatives and keep our "plug-in" reconstruction module updated.  By harnessing the power of advanced algorithms, we aim to achieve a significant breakthrough in the quality of 3D reconstructions, opening new doors for applications that demand precise and comprehensive 3D models.
}


\bibliographystyle{ACM-Reference-Format}
\bibliography{sample-bibliography}

\appendix
\label{sec:appendix}
\begin{figure*}
 \centering
 \begin{minipage}{\textwidth}
 \centering
 \includegraphics[width= 0.85\textwidth]{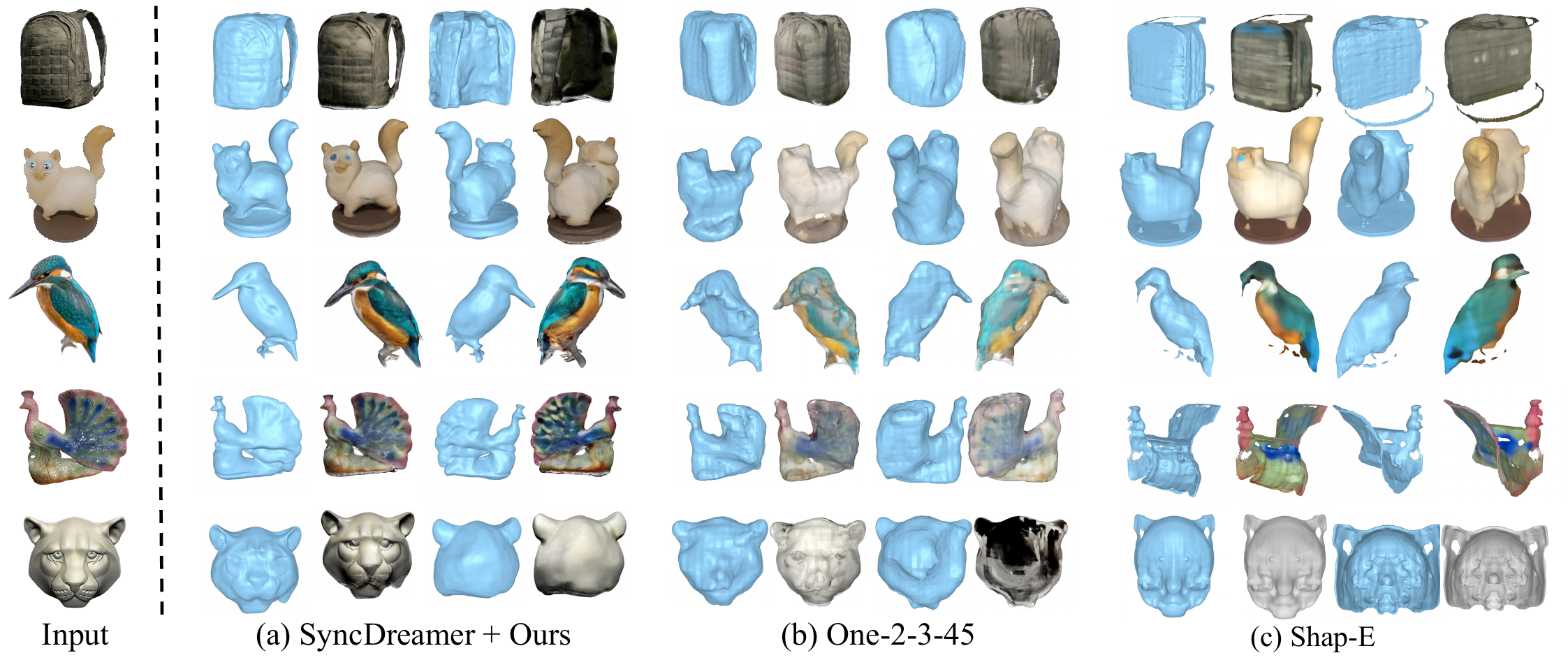}
 \captionof{figure}{Based on a single-view input, we reconstruct the object using images generated by SyncDreamer \cite{liu2023syncdreamer} and compare our results with One-2-3-45 \cite{liu2023one} and Shap-E \cite{jun2023shap}. Our outcomes demonstrate significant advantages in both texture and geometry.}
 \label{fig:compare}
 \end{minipage}
\end{figure*}

\begin{figure*}[!htbp] 
 \centering
 \begin{minipage}{\textwidth}
 \centering
 \includegraphics[width=  0.9\textwidth]{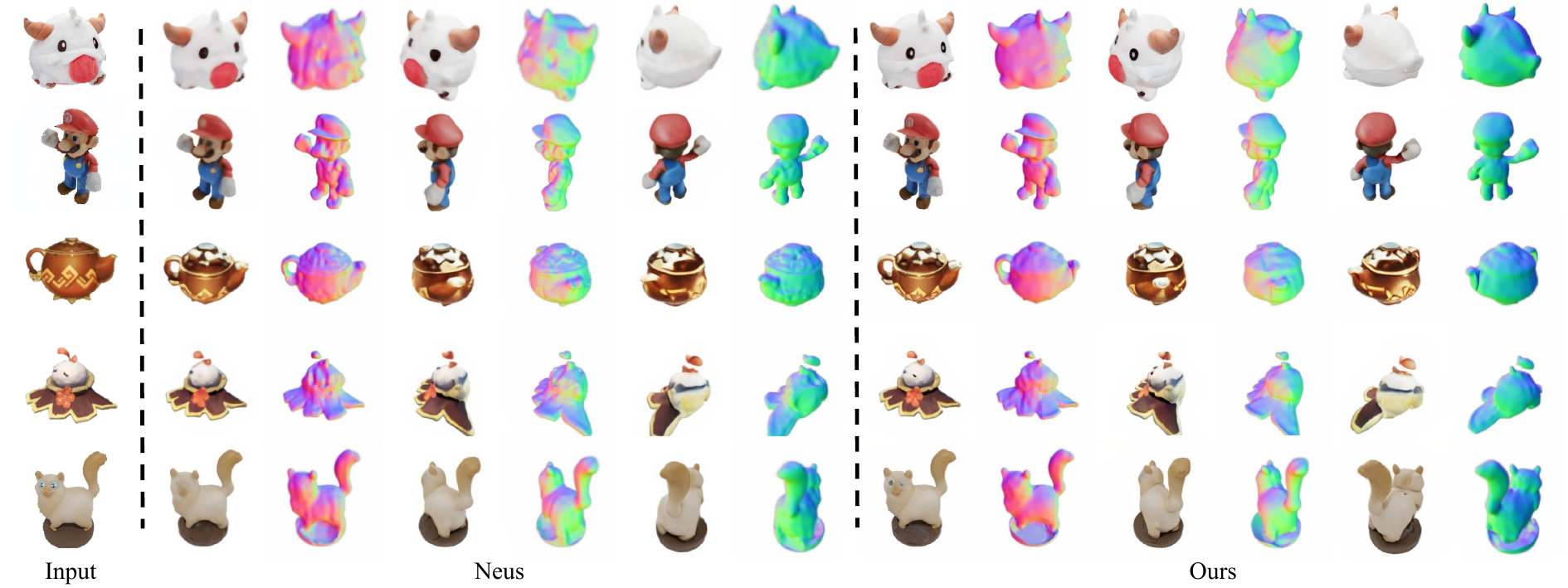}
 \captionof{figure}{Rendering results of our method on SyncDreamer\cite{liu2023syncdreamer}-generated images.}
 \label{fig:MVDream}
 \end{minipage}
\end{figure*}


\begin{figure*}[!htbp] 
 \centering
 \begin{minipage}{\textwidth}
 \centering
 \includegraphics[width=  0.9\textwidth]{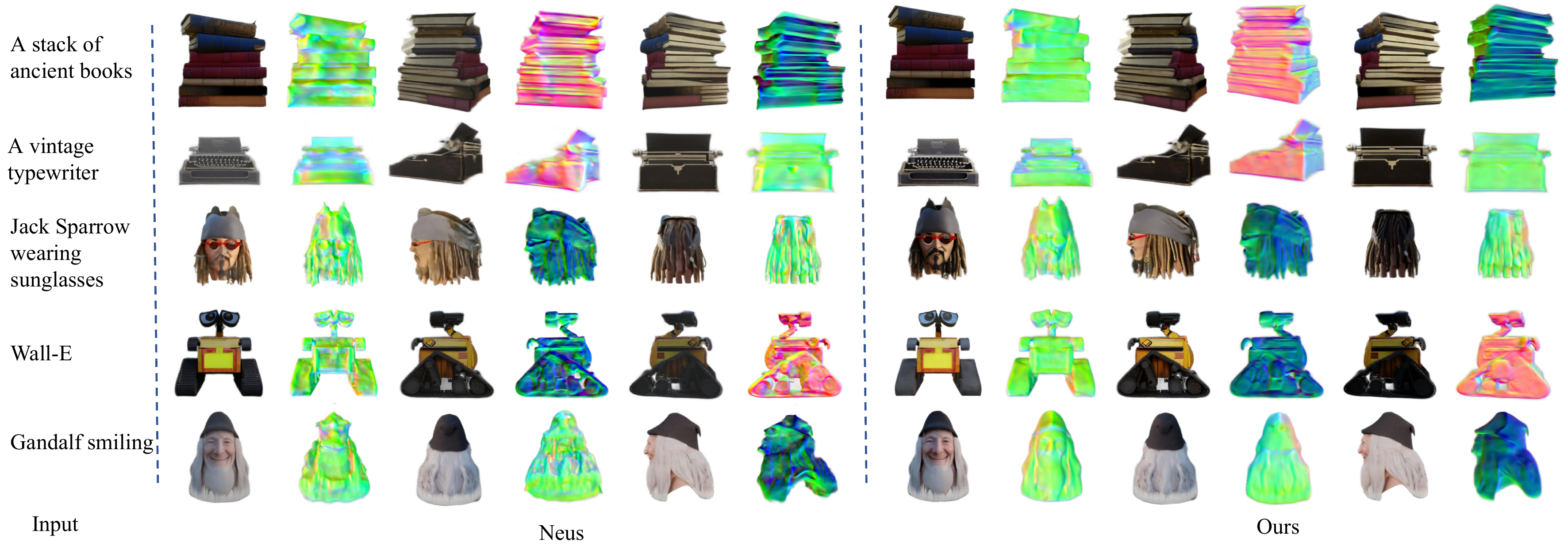}
 \captionof{figure}{Rendering results of our method on text-generated (MVDream\cite{shi2023mvdream}) images.}
 \label{fig:MVDream}
 \end{minipage}
\end{figure*}

\begin{figure*}[ht]
 \centering
 \begin{minipage}{\textwidth}
 \centering
 \includegraphics[width= \textwidth]{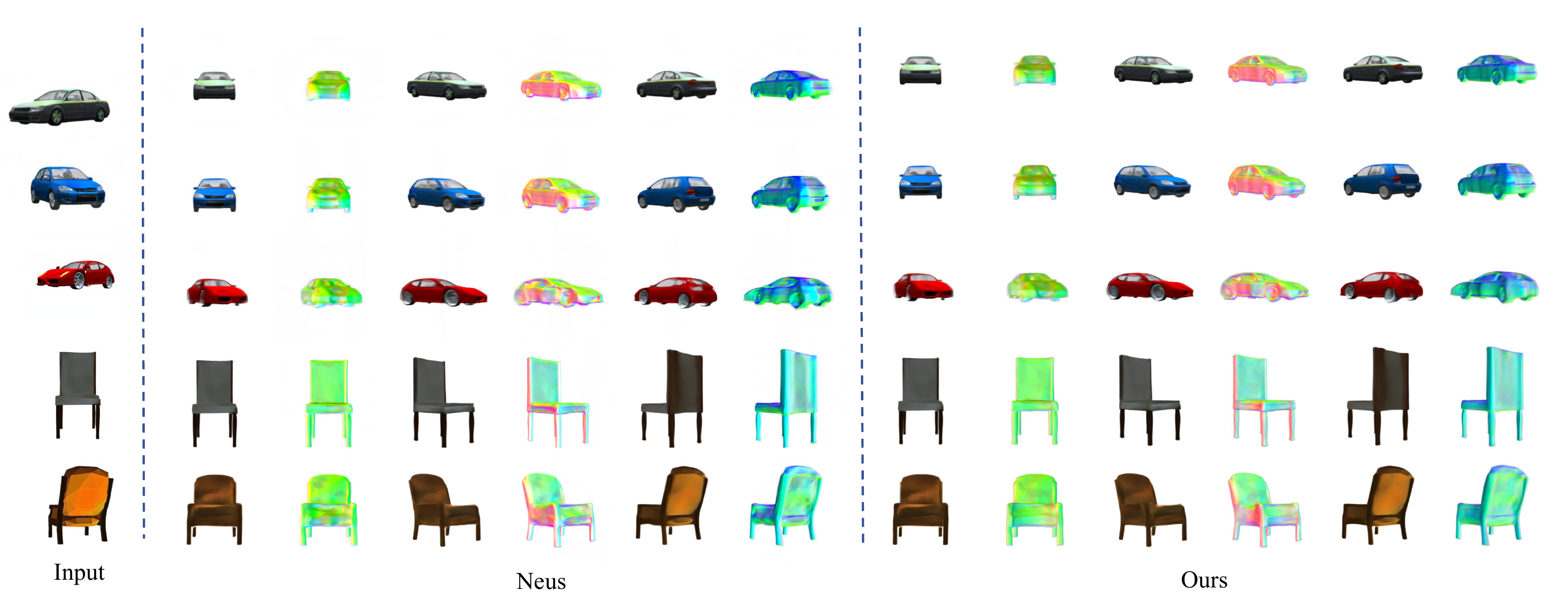}
 \captionof{figure}{Rendering results of our method on GFLA\cite{ren2020deep}-generated images.}
 \label{fig:gfla}
 \end{minipage}
\end{figure*}

\begin{figure*}[ht]
 \centering
 \begin{minipage}{\textwidth}
 \centering
 \includegraphics[width= \textwidth]{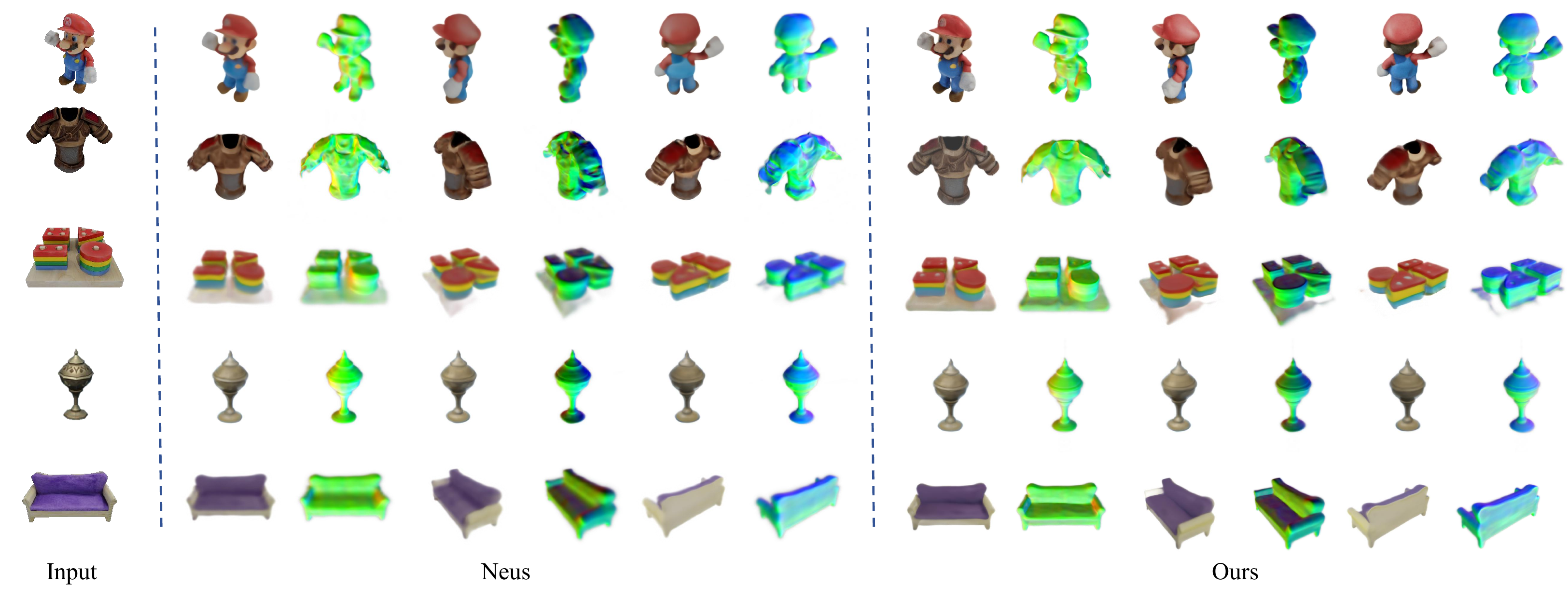}
 \captionof{figure}{Rendering results of our method on zero123 \cite{liu2023zero}-generated images}
 \label{fig:zero123}
 \end{minipage}
\end{figure*}

\begin{figure*}[ht]
 \centering
 \begin{minipage}{\textwidth}
 \centering
 \includegraphics[width= \textwidth]{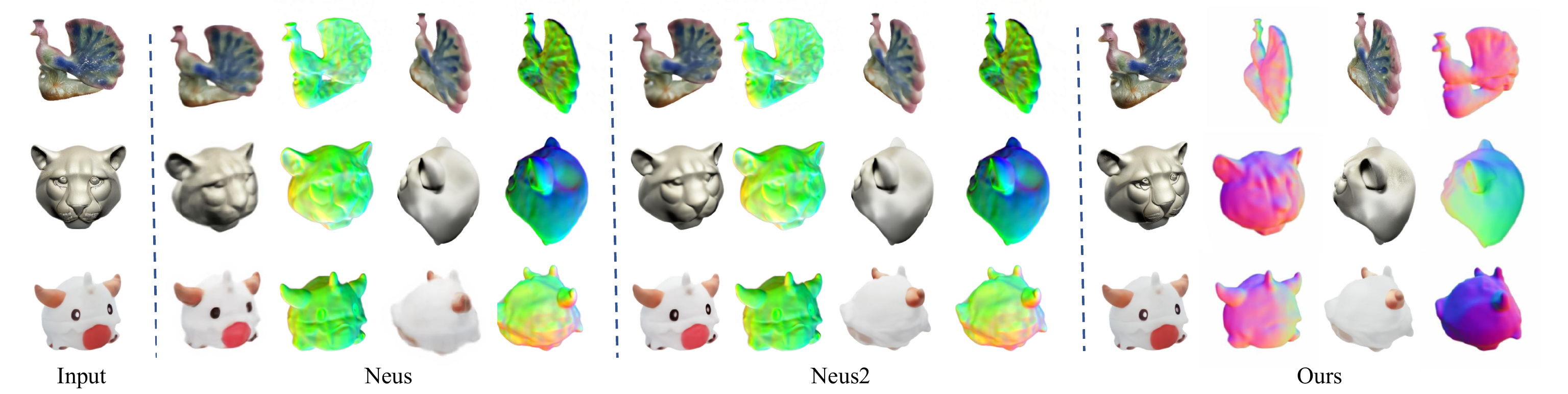}
 \captionof{figure}{Visual comparison of reconstruction using Neus, Neus2 and our method comparison on SyncDreamer-generated images. }
 \label{fig:nerf}
 \end{minipage}
\end{figure*}


\end{document}